%% file: main.tex
\documentclass[sigconf]{acmart}

\AtBeginDocument{%
  \providecommand\BibTeX{{%
    \normalfont B\kern-0.5em{\scshape i\kern-0.25em b}\kern-0.8em\TeX}}}

\usepackage{xspace}
\usepackage{tikz}
\usepackage{subfigure}
\usepackage{float}
\usepackage{multirow}
\usepackage{caption} 
\newcommand*\circled[1]{\tikz[baseline=(char.base)]{
            \node[shape=circle,draw,inner sep=0.1pt] (char) {#1};}}

\newcommand{\lehdc}{LeHDC\xspace}
\newcommand{\ouralg}{LeHDC\xspace}

\setcopyright{none}
\renewcommand\footnotetextcopyrightpermission[1]{}
\begin{document}

\title{\lehdc: Learning-Based Hyperdimensional Computing Classifier}

\author{Shijin Duan}
\email{duan.s@northeastern.edu}
\affiliation{
 \institution{Northeastern University}
 \city{Boston}
 \state{MA}
 \country{USA}
}

\author{Yejia Liu}
\email{yliu807@ucr.edu}
\affiliation{
 \institution{UC Riverside}
 \city{Riverside}
 \state{CA}
 \country{USA}
}

\author{Shaolei Ren}
\email{sren@ece.ucr.edu}
\affiliation{
 \institution{UC Riverside}
 \city{Riverside}
 \state{CA}
 \country{USA}
}

\author{Xiaolin Xu}
\email{x.xu@northeastern.edu}
\affiliation{
 \institution{Northeastern University}
 \city{Boston}
 \state{MA}
 \country{USA}
}

\begin{abstract}
Thanks to the tiny storage and efficient execution, hyperdimensional Computing (HDC) is emerging as a lightweight learning framework on resource-constrained hardware. Nonetheless, the existing HDC training relies on various heuristic methods, significantly limiting their inference accuracy. In this paper, we propose a new  HDC framework, called \lehdc, which leverages a principled learning approach to improve the model accuracy. Concretely, \lehdc\ maps the existing HDC framework into an equivalent Binary Neural Network architecture, and employs a corresponding training strategy to minimize the training loss. Experimental validation shows that \lehdc\ outperforms previous HDC training strategies and can improve on average the inference accuracy over 15\% compared to the baseline HDC.
\end{abstract}

\maketitle

\section{Introduction}
\label{sec:introduction}
Brain-inspired hyperdimensional computing (HDC) is raised to represent samples by projecting them to extremely high-dimensional vectors, i.e., \textit{hypervector} \cite{kanerva2009hyperdimensional}. As an emerging method, HDC is a promising alternative to conventional machine learning models like deep neural networks (DNNs), with less storage usage and higher efficiency. Although HDC is not 
meant to replace DNNs on all complex classification tasks, it indeed shows impressive performance on lightweight tasks and fits well in highly resource-limited Internet-of-Things (IoTs) devices. Given these characteristics, the studies on HDC have been quickly proliferating, including energy efficiency improvement \cite{imani2019quanthd, imani2019searchd} and applications on tiny devices \cite{imani2018hdna, thapa2021spamhd}. Meanwhile, HDC models have also been adopted on various acceleration platforms, such as FPGA \cite{imani2019quanthd}, GPU \cite{kim2020geniehd}, and in-memory computing \cite{karunaratne2020memory}, thanks to their high parallelism capacity.

Depending on the format of hypervectors, HDC can be broadly divided into binary HDC and non-binary HDC: the hypervectors and computations in binary HDC are all binarized, while binarization
is not used in non-binary HDC. Naturally, non-binary HDC contains richer information expression on hypervectors, but it also costs more computing resources than binary HDC. On the other hand, binary HDC consumes lower energy and resources and is more friendly to hardware implementation. 
More recently, some heuristic approaches, such as retraining (i.e., fine-tune the class hypervectors after initial training) \cite{imani2019quanthd}, have been proposed to improve the inference accuracy, making binary HDC achieve a competitive accuracy performance compared to its non-binary counterpart. 
Nonetheless, the initial training process for an HDC model, binary and non-binary, still heavily depends on a simple strategy of \emph{averaging} the sample hypervectors to obtain class hypervectors. In other words, there have been no principled approaches to  optimally train an HDC model and rigorously learn the  class hypervectors that provide the best possible accuracy performance for HDC.

In this paper, we demonstrate for the first time that a binary HDC classifier is equivalent to a wide single-layer binary neural network (BNN).\footnote{Our result also applies to non-binary HDC models by changing the BNN to a wide single-layer neural newtork with non-binary weights.} Specifically, the binary weights in the BNN can be viewed as the class hypervectors in binary HDC, and the Hamming distance between the encoded hypervector and the class hypervectors in binary HDC can be linearly transformed into multiplication of the encoded hypervector and the binary weights. By viewing the class hypervector training process from the BNN perspective, we reveal the key limitations  of the 
current HDC training strategy: heavily relying on heuristic approaches to search for class hypervectors. To address this limitation, we propose a learning-based HDC training strategy, namely \lehdc. Specifically,  \ouralg 
takes the sample hypervector as input,
assigns \textit{one-hot} labels,
and optimizes the BNN weights 
in the training process. The binary weights in the BNN are trained with state-of-the-art
learning algorithms, and consequently, these binary weights can be directly converted to class hypervectors for the binary HDC classification. The binary HDC with obtained class hypervectors can achieve significantly higher accuracy than the current retraining strategies. Importantly,
\lehdc introduces a completely new training process, but
does not modify the encoding or inference processes used in the existing HDC. As a result,
\ouralg can be integrated into any existing HDC framework to improve the accuracy performance,
yet without any extra resource or execution overhead during the inference.

The main contributions of this work are as follows:
\begin{itemize}
    \item We transform the existing binary HDC classifier into an equivalent BNN,
    and then reveal the key limitations in the current HDC training process that limit them from obtaining the optimal class hypervectors and achieving the best accuracy.
    \item We propose a new training strategy, \lehdc, on binary HDC classification, by leveraging state-of-the-art BNN learning algorithms. To the best of our knowledge, this is the first work using learning-based methods to train HDC classifications in a principled manner.
    \item We show empirically that
    \ouralg can significantly outperform current HDC models and provide over $15\%$ accuracy improvement against the baseline HDC, while introducing zero resource and time overhead during inference.
\end{itemize}

Our paper organization is: In Sec. \ref{sec:background}, we briefly discuss the HDC classification and current training strategies. Sec. \ref{sec:defects} reveals the limitations
of the current HDC training process and equivalently expresses the HDC model in a wide single-layer BNN structure. Sec. \ref{sec:proposal} illustrates the proposed training strategy in \lehdc, and Sec. \ref{sec:validation} presents the feasibility of our strategy and compares the performance with other training strategies. The conclusion and discussion on our work are addressed in Sec. \ref{sec:conclusion}.

\section{HDC Classification Tasks}
\label{sec:background}

Binary HDC has been emerging as a novel paradigm that represent attributes with hyperdimensional bi-polar vectors $\{1, -1\}^D$ \cite{kanerva2009hyperdimensional}. For a specific sample $\textbf{F} = \{f_1, f_2, ..., f_N\}$ where $f_i$ is the value of the $i$-th feature for $i=1,\cdots,N$, its feature positions and feature values are represented by randomly generated hypervectors, whose dimension (e.g., $D = 10,000$) is much larger than the number of features/values. In
a typical HDC model \cite{imani2019quanthd}, feature position hypervectors ($\mathcal{F}$) are  orthogonal to identify an individual feature, i.e., the normalized Hamming distance $Hamm(\mathcal{F}_i, \mathcal{F}_j) \approx 0.5, i,j\in \{1,2,...,N\}$. Differently, feature value hypervectors ($\mathcal{V}$) are correlated to reflect the correlations in real values, i.e., $Hamm(\mathcal{V}_{f_i}, \mathcal{V}_{f_j}) \propto\frac{|f_i - f_j|}{max - min}$, where $f_i$ and $f_j$ are two samples in the value range, $f_i,f_j\in [min, max]$.

\subsection{Binary HDC}
Binary HDC costs much lower power and computational resources, 
and is also the mainstream HDC framework \cite{imani2019quanthd}. By binding feature position hypervectors and value hypervectors, a sample can be described as a new hypervector ($\mathcal{H}\in\{1,-1\}^D$):
\begin{equation}
    \mathcal{H} = sgn\left(\sum_{i=1}^N \mathcal{F}_i\circ \mathcal{V}_{f_i}\right)
\label{eq:encoding}
\end{equation}
where this sample has $N$ features, and $\circ$ denotes the Hadamard product to multiply two hypervectors in element-wise. $sgn(\cdot)$ is the sign function to binarize the sum of hypervectors; here we assume $sgn(0)$ is randomly assigned with 1 or -1. In general,
an HDC classifier can use record-based or $N$-gram-based encoders \cite{ge2020classification}. While our training approach applies to any encoding methods (including
advanced ones \cite{HDC_Cornell_arXiv_2022} based on sophisticated feature extractions), \footnote{Per the DAC’22 policy, we are not allowed to make significant non-editorial changes to papers once accepted. Thus, as Ref.~\cite{HDC_Cornell_arXiv_2022} (arXiv date 2/10/2022) was not
available or cited at the time of our DAC’22 submission
on 11/22/2021, it will not be included in the final camera-ready version of this paper.} for a concrete case study, we adopt the commonly-used \textit{record-based encoding} which, shown in Eq.~\ref{eq:encoding}, has higher accuracy than the $N$-gram-based method for many applications \cite{ge2020classification}.
Note that \ouralg does not modify the encoding process, and hence can work with any encoders.

\begin{figure}[htbp]
\centering
\includegraphics[width=\linewidth]{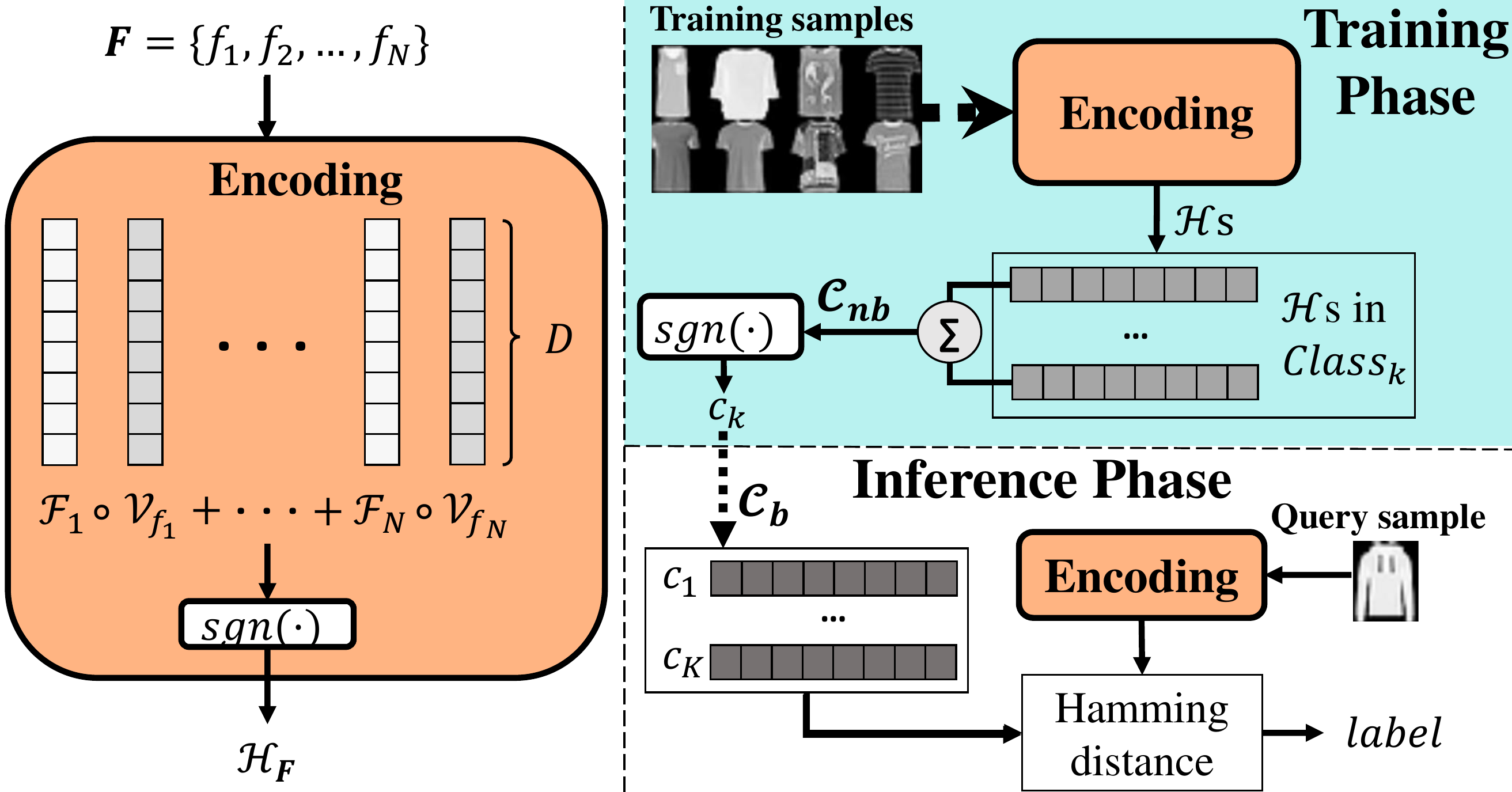}
\caption{Binary HDC classification framework.}
\label{fig:binaryHDC}
\end{figure}

\textbf{Training.} The basic training strategy in HDC is to simply accumulate all the samples belonging to that class, in order to obtain the class hypervectors $\mathcal{C}$:
\begin{equation}
    c_k = sgn\left(\sum_{\mathcal{H}\in \Omega_k} \mathcal{H}\right)
\label{eq:initial_training}
\end{equation}
where $c_k$ denotes the $k$-th class hypervector in $\mathcal{C}$, and $\Omega_k$ is the set of sample hypervectors belonging to class $k$. 

\textbf{Inference.} 
A query sample is first encoded using Eq. \ref{eq:encoding}. Then, the similarities,
measured in terms of the Hamming distance between the query hypervector and class hypervectors in $\mathcal{C}$ are calculated. The most similar one, i.e., the class with the lowest Hamming distance, is labeled as the predicted class. For the ease of understanding the HDC flow, we show the scheme of binary HDC classification in Fig. \ref{fig:binaryHDC}. This procedure is similar to the nearest centroid classification in machine learning \cite{levner2005feature}, which searches for an optimal centroid for each class.

\subsection{Training Enhancement}
Various training strategies have been proposed
to increase accuracy. Here, we introduce a state-of-the-art approach: retraining \cite{imani2019quanthd}.

\begin{figure}[htbp]
\centering
\includegraphics[width=0.8\linewidth]{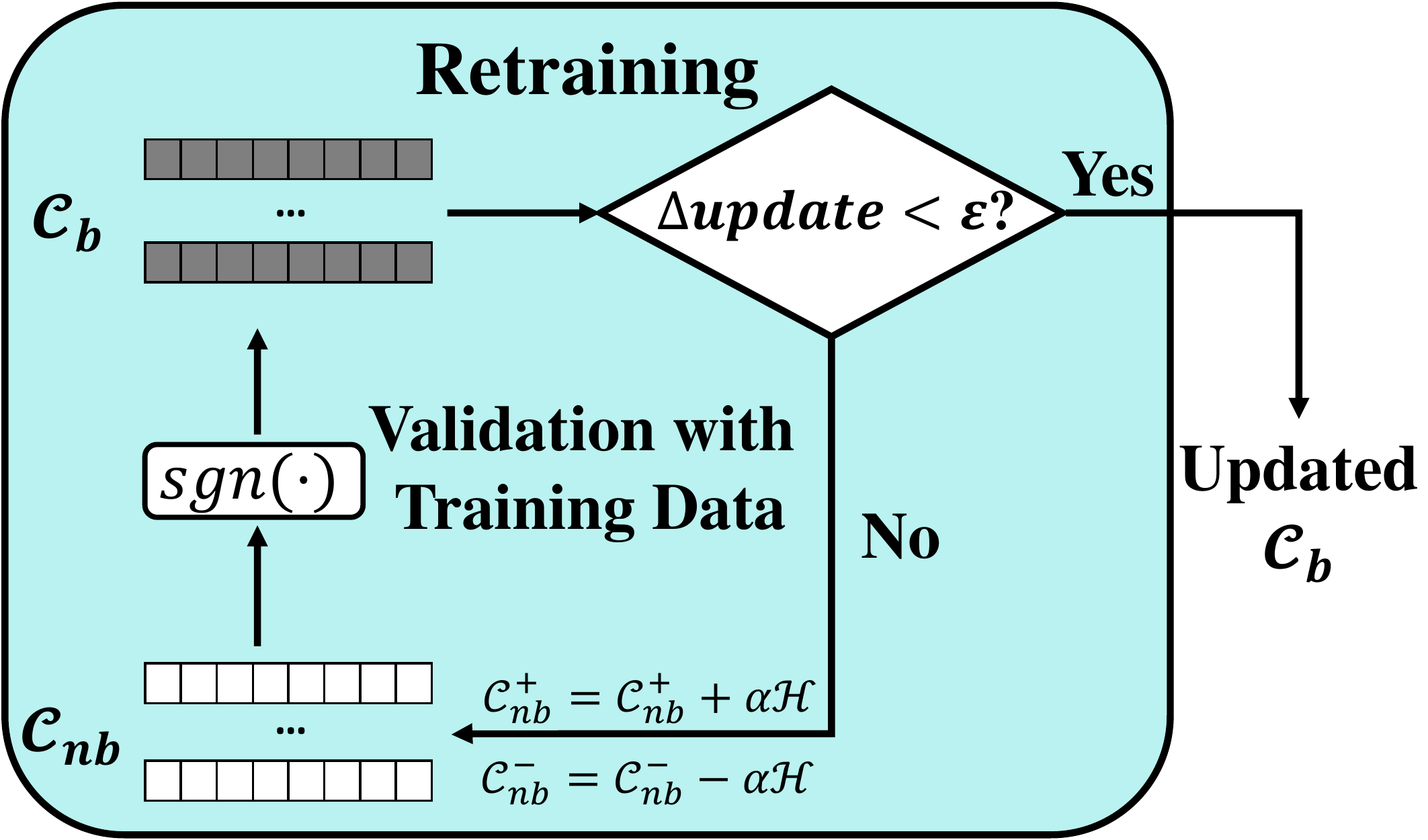}
\caption{Retraining strategy to adjust class hypervectors against misclassified samples.}
\label{fig:retraining_model}
\end{figure}

Eq. \ref{eq:initial_training} gives the initial training results for class hypervectors. The retraining strategy \cite{imani2019quanthd}  further fine-tunes the initial class hypervectors, in which both non-binary and binary class hypervectors are used for the training. Specifically, the binary class hypervectors are utilized for validation and non-binary ones are used for updating. As shown in Fig. \ref{fig:retraining_model}, in each retraining iteration, training samples are classified based on the current binary class hypervectors. If a training sample is misclassified, then the non-binary class hypervectors will be updated regarding to the encoded hypervector ($\mathcal{H}$):
\begin{equation}
\begin{split}
    c_{nb}^+ &= c_{nb}^+ + \alpha \mathcal{H}\\
    c_{nb}^- &= c_{nb}^- - \alpha \mathcal{H}
\end{split}
\label{eq:retraining}
\end{equation}
where $c_{nb}^+$ and $c_{nb}^-$ are the correct and misclassified non-binary class hypervectors, respectively, and $\alpha$ is referred to as the learning rate. This retraining step intends to increase the influence of misclassified sample on the correct class while reducing it on the misclassified class. The retraining stops when the updating on class hypervectors is negligible. 

In addition, an ensemble approach (e.g., multi-model HDC where multiple models collectively classify each sample \cite{imani2019searchd}) can also increase the accuracy, but the storage size will grow when the number of ensembled HDC models increases.

\section{Inner Mechanism of HDC Classifiers}
\label{sec:defects}
In this section, we demonstrate the equivalence
of a binary HDC classifier to a corresponding BNN
for inference. Importantly, we highlight that
 the existing training strategies for HDC,
are mostly heuristics and hence not optimal.

\subsection{From Binary HDC to BNN}
Considering a binary HDC classifier, we denote the input feature of a sample as $x\in\mathbb{R}^N$. The encoder in binary HDC will transfer the real-valued input feature $x$ to a binary hypervectors: $En(x): \mathbb{R}^{N} \mapsto \{-1, 1\}^{D}$, where $D$ stands for the dimension of each hypervector, i.e., projecting the sample input from a low dimensional space to a much higher dimension. Assuming there are $K$ classes for this HDC classifier, a trained class hypervector set is $\mathcal{C} = \{c_1, c_2, ..., c_K\} \in \{-1, 1\}^{D\times K}$, where $c_k$ is the $k$-th class hypervector. The predicted label for the sample $x$ is 
\begin{equation}
    k^\star = \underset{k}{argmin}\ Hamm(En(x), c_k),
\end{equation}
where $Hamm(\mathcal{H}_1, \mathcal{H}_2) = \frac{|\mathcal{H}_1\neq \mathcal{H}_2|}{D}$ represents the normalized Hamming distance operator between any two hypervectors $\mathcal{H}_1$ and $\mathcal{H}_2$, and $|\mathcal{H}_1\neq \mathcal{H}_2|$ denotes the number of different bits in $\mathcal{H}_1$ and $\mathcal{H}_2$. A key property is that the hamming distance can be equivalently projected to the cosine similarity: $cosine(\mathcal{H}_1, \mathcal{H}_2) = 1 - 2\cdot Hamm(\mathcal{H}_1, \mathcal{H}_2)$.

To see this point more concretely, we  write the cosine similarity of
 two binary hypervectors 
$\mathcal{H}_1$ and $\mathcal{H}_2$ as
\begin{equation}
    \begin{split}
    cosine(\mathcal{H}_1, \mathcal{H}_2)  = \frac{\mathcal{H}_1 ^T \mathcal{H}_2}{\|\mathcal{H}_1\|\ \|\mathcal{H}_2\|}
    \end{split}
\end{equation}
where $\|\mathcal{H}_1\|$ and $\|\mathcal{H}_2\|$ denote the $l_2$ norms of 
$\mathcal{H}_1$ and $\mathcal{H}_2$, respectively. Due to the bipolar values $\{1, -1\}$ in hypervectors, we have $\mathcal{H}_1^T \mathcal{H}_2 = (|\mathcal{H}_1 = \mathcal{H}_2| - |\mathcal{H}_1 \neq \mathcal{H}_2|)$. Plus the fact that $\|\mathcal{H}_1\|\ \|\mathcal{H}_2\| = D$ and $(|\mathcal{H}_1\neq \mathcal{H}_2| + |\mathcal{H}_1= \mathcal{H}_2|) = D$, we can conclude the equivalence between the Hamming distance and cosine similarity.

Therefore, the predicted label $k^\star$ can be equivalently represented as
\begin{equation}\label{eq:equal_to_BNN}
\begin{split}
    k^\star &=\underset{k}{argmin}\ Hamm(En(x), c_k)\\
    &= \underset{k}{argmax}\ cosine(En(x), c_k)\\
    &= \underset{k}{argmax}\ En(x)^T c_k
\end{split}
\end{equation}
\textbf{Remark.} From Eq. \ref{eq:equal_to_BNN}, we argue that the computation $\ En(x)^T c_k$ is actually the same as forward propagation in a BNN. Specifically, as illustrated in Fig.~\ref{fig:bnnNetwork}, the single-layer BNN takes $En(x)$ as its input and has $K$ output neurons that represent $K$ classes. The binary connection weights between the input $En(x)\in \{-1,1\}^D$ and the $k$-th output neuron is $c_k\in\{-1,1\}^D$, while the $k$-th output is $\ En(x)^T c_k$ and non-binary. 

While binary HDC is the mainstream choice for HDC classifiers, our analysis also applies to non-binary HDC, where the encoded hypervector $En(x)$ and class hypervector $c_k$ can take both non-binary values. In this case, cosine similarity is directly used as the measure between $En(x)$ and $c_k$ for classification, and a non-binary HDC can be equivalently viewed as a simple single-layer neural network (i.e., perceptron).

\subsection{Limitations in Current HDC Training}

By establishing the equivalence between
a binary HDC model and a BNN, we can see
that the HDC training process (i.e., finding
 class hypervectors $\{c_1,\cdots,c_K\}$) is essentially
 the same as training the BNN weights $c_1,\cdots,c_K$. 
 
In the basic HDC training process,
each class hypervector $c_k$ is obtained by simply
averaging the sample hypervectors $En(x)$ of
all the samples belonging to that class.
Clearly, this naive approach does not optimize
the BNN weight $c_k$ at all.
Next, we also highlight the key limitations in the
state-of-the-art retraining strategy.

\textbf{(1)} \textit{Retraining only updates non-binary class hypervectors that correspond to the misclassified class and the true class, while other class hypervectors stay the unchanged.} Intuitively, when a training sample $x$ is misclassified, the retraining step can partially mitigate the impact of $En(x)$ on the misclassified class hypervector while enhancing its impact on the true one; when a training sample is correctly classified, no action is taken. However, this strategy neglects two scenarios that could occur during the retraining phase: \circled{1} If a training sample $x$ is misclassified, while there are multiple wrong labels with high similarity, only the class hypervector corresponding to the wrong label with the highest similarity is updated. Hence, other wrong class hypervectors can only be updated in the future iterations, or even never get updated. \circled{2} If a training sample $x$ is correctly classified, even though the correct label has a slightly higher similarity than other classes (i.e., the other class hypervectors also have high similarities to this sample), no class hypervectors are updated. In  this case, albeit correctly classified, this sample is very close to the classification border. If more samples of this kind occur during the training process, \textit{over-fitting} is likely to happen, thus weakening the generalization of the HDC model. In contrast, there are various mechanisms, such as dropout and regularization, which mitigate the over-fitting in BNNs. Thus, by  training the equivalent BNN, we can systematically improve the testing accuracy of HDC models.
  
\textbf{(2)} \textit{All the weights along the updating class hypervector have to be updated with a fixed step size.} In the retraining strategy \cite{imani2019quanthd}, the updating scale is only determined by the encoded hypervector $En(x)$ and a fixed learning rate $\alpha$, as shown in Eq.\ref{eq:retraining}. On the other hand, the general updating rule for a (single-layer) DNN is:
\begin{equation}
    c^{t+1}_{k,j} = c^t_{k,j} - \alpha \frac{\partial \mathcal{L}}{\partial c_{k,j}} x
\label{eq:updating}
\end{equation}
where $c_{k,j}^t$ is one weight parameter for iteration $t$, $\mathcal{L}$ denotes the loss function, $\alpha$ is the learning rate, and $x$ stands for the sample input equivalent to $En(x)$ in the HDC model. By comparing Eq.\ref{eq:retraining} and Eq.\ref{eq:updating}, we can directly observe that the derivative of the error $\frac{\partial \mathcal{L}}{\partial c_{i,j}}$ is overlooked in the retraining strategy. For this single-layer network, the derivative of loss function can reflect the similarity between the input $En(x)$ and the class hypervectors $c_k, k\in\{1,2,...,K\}$. However, the retraining strategy indeed does not consider the similarity during the updating. As an improved version, adaptive learning rate is proposed in \cite{imani2019adapthd}, but the adaptability is still determined on the validation error rate or the difference between the similarities of $cosine(En(x),c_{correct})$ and $cosine(En(x),c_{wrong})$, not the similarities themselves on all class hypervectors. Consequently, the state-of-the-art retraining strategy will converge very slowly due to updating with incomplete information. 

\subsection{Case Study}
We now practically demonstrate how thes limitations can affect HDC training. For the case study, we use Fashion-MNIST dataset \cite{xiao2017/online} to illustrate the inner mechanism of HDC learning. Fashion-MNIST consists of $L=60,000$ training images which are classified into $K=10$ classes, and we set the hypervector dimension as $D=10,000$.

\begin{figure}[htbp]
\centering
	\subfigure[Training trajectory]{
	    \centering
		\includegraphics[width=0.48\linewidth]{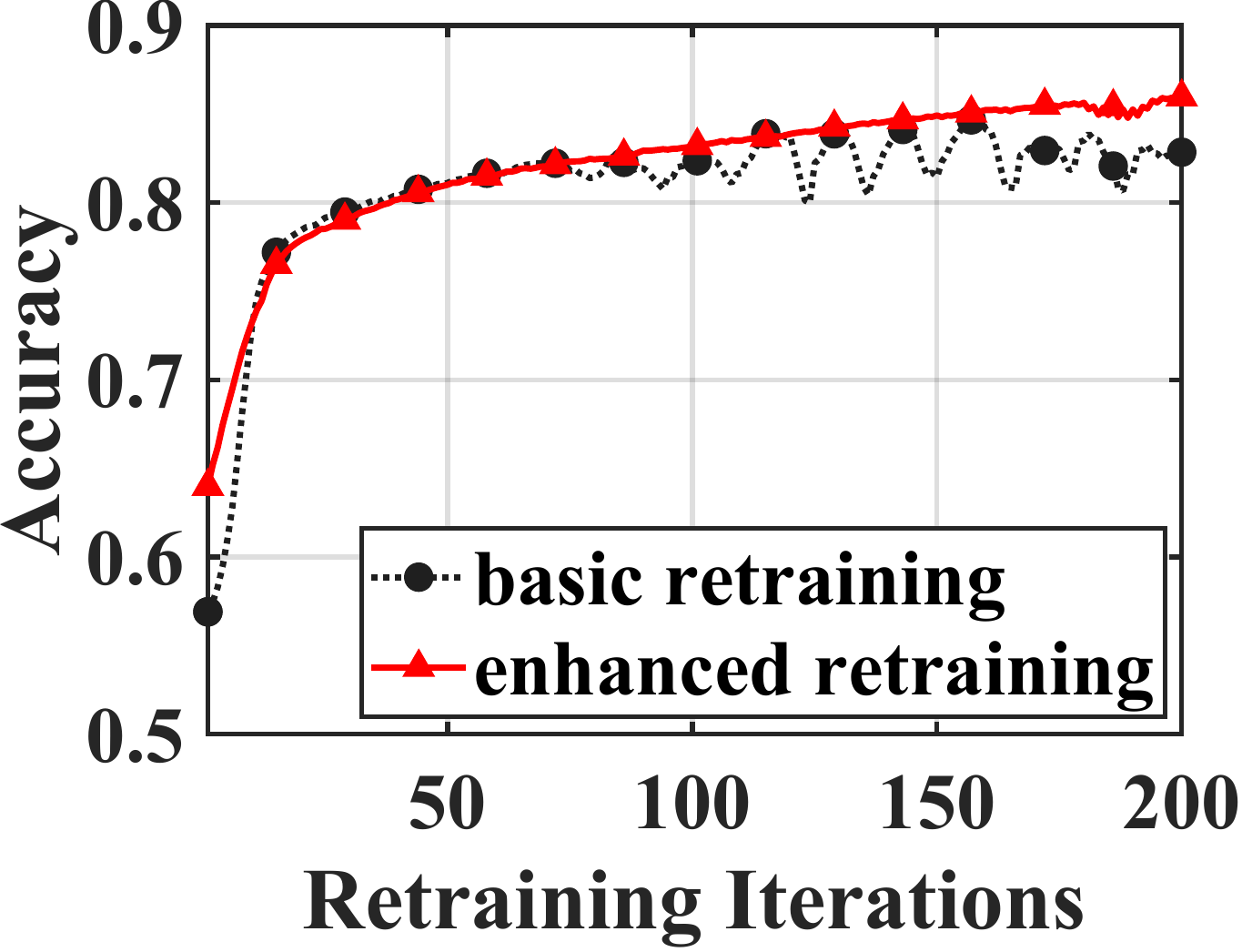}
		\label{fig:retraining_training}}
		~~
	\subfigure[Testing trajectory]{
	    \centering
		\includegraphics[width=0.48\linewidth]{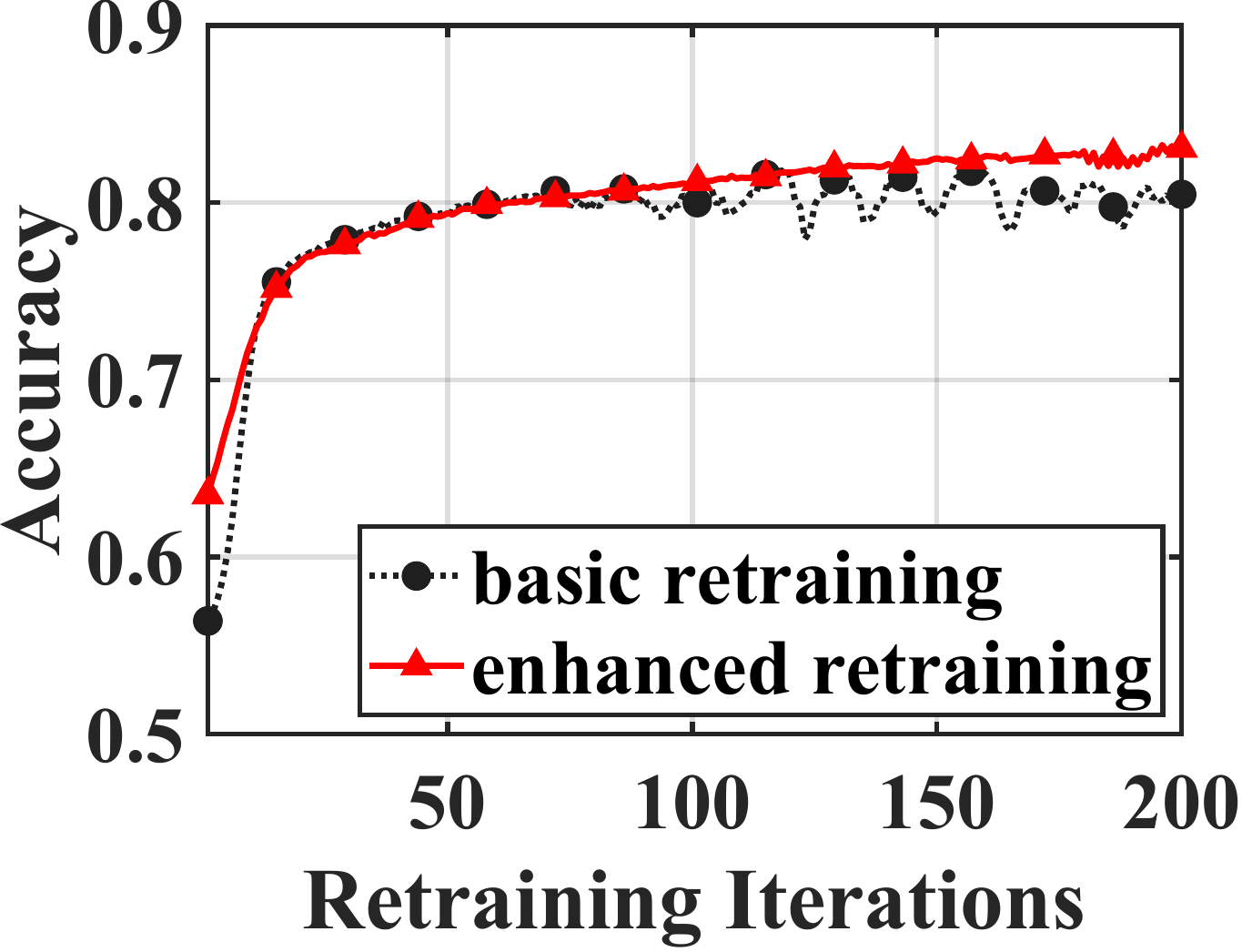}
		\label{fig:retraining_testing}}
\caption{Iteration comparison on the basic \cite{imani2019quanthd} and the enhanced retraining strategy. The enhanced method adds similarity consideration and multiple updates during retraining.}
\label{fig:comparison}
\end{figure}

In Fig. \ref{fig:comparison}, we compare the performance of enhanced retraining  against that of the default retraining. We  make modifications on top of the existing retraining strategy to enhance the retraining process. Specifically, once a training sample is misclassified, all the class hypervectors that have higher similarities than the correct class hypervector will be updated, instead of only the one with the highest similarity.  During the updating, we add the similarity influence. We specify that the ideal Hamming distance between the training sample and the correct/wrong class hypervector is 0/0.5. Then, we calculate the difference between the Hamming distance and the ideal one, and use it as a scaling factor for updating a class hypervector during retraining. This is equivalent to Eq.~\ref{eq:updating} when the loss function is the squared error. 

The result shows that, for both the training and testing procedures, the enhanced retraining  strategy can start with and converge at a higher accuracy, affirming that the discussed limitations indeed limit the retraining strategy. On the other hand, the basic retraining strategy starts to oscillate after the initial convergence. In contrast, the enhanced retraining can make the training/testing procedure more stable, due to the introduction of similarity metric for scaling the updating steps. Nonetheless, the enhancements we make are still heuristic, lacking  principled guidance to optimize the class hypervectors in HDC.

\section{\ouralg: the Learning-Based HDC}
\label{sec:proposal}
We now present \lehdc as an alternative and principled approach to train
the class hypervectors in an HDC classifier. Based on the discovered equivalence
between an HDC model and a single-layer BNN, \ouralg leverages state-of-the-art
principled learning algorithms to train the BNN weights.

\begin{figure}[htbp]
\centering
\includegraphics[width=0.8\linewidth]{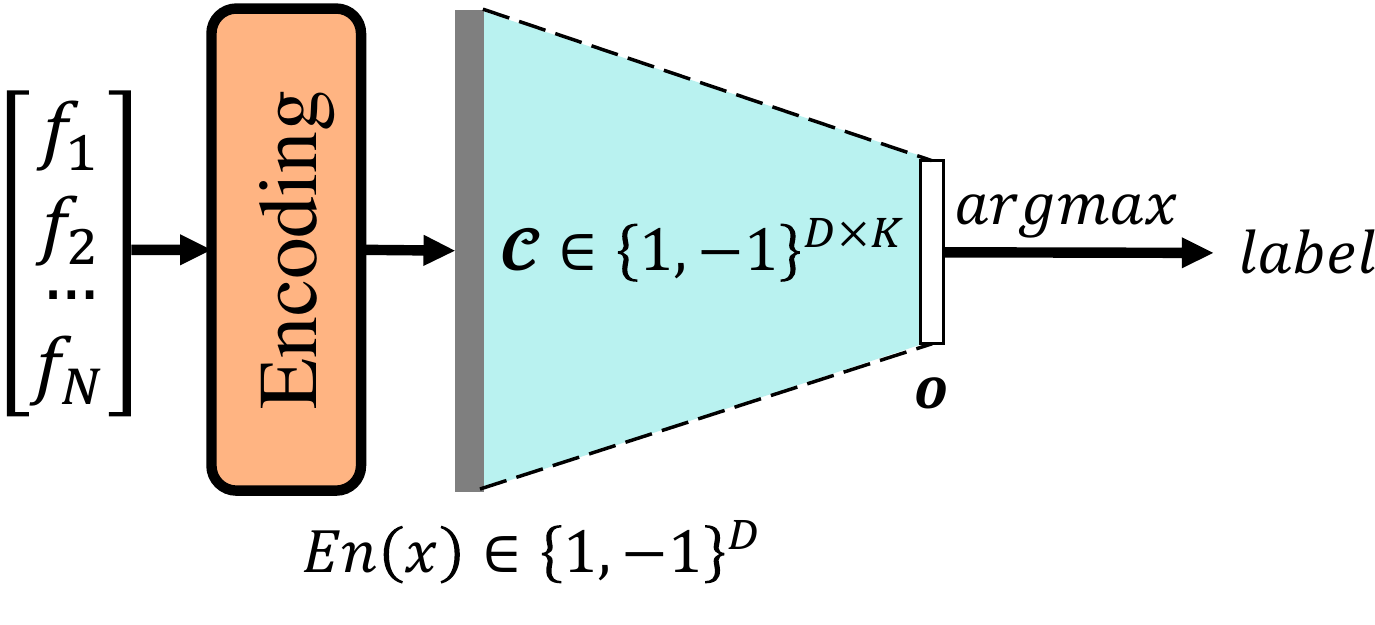}
\caption{The equivalent BNN model for binary HDC.}
\label{fig:bnnNetwork}
\end{figure}

Compared to non-binary neural networks, BNN is more challenging to train, as the weights and output values are all binary. For instance, a large learning rate may successfully flip the binary weights but introduce severe oscillation at the same time; while a small learning rate may not be powerful enough to flip binary bits, resulting in the updating trapped in local optima. A BNN model for the binary HDC learning is demonstrated in Fig. \ref{fig:bnnNetwork}. Here, $En(x)$ is an encoded sample hypervector and the input to the BNN, $\mathcal{C}$ represents the class hypervectors and are the BNN weights, and output $\textbf{o}$ is $(En(x)^Tc_1,\cdots, En(x)^Tc_K)$ and equivalently measures the similarities between the input and each class. 
In this paper, we adopt the state-of-the-art BNN training strategy in \cite{liu2021adam}, and propose the following approach to obtaining the optimal class hypervectors. 

Unlike other BNN models, our single-layer BNN corresponding to the binary HDC model does not require the binary activation function at each output neuron, since the non-binary BNN outputs (i.e., $\ En(x)^T c_k$, for $k=1,\cdots,K$) are directly used to determine the classification result. For the binary weights $\mathcal{C}\in \{-1, 1\}^{D\times K}$ (i.e., class hypervectors), both binary ($\mathcal{C}$) and non-binary ($\mathcal{C}_{nb}$) forms are stored during training. The non-binary hypervectors are utilized to accumulate small gradients, and they are updated during the back propagation. The binary hypervectors are utilized for feed-forward and updated after each iteration as:
\begin{equation}
    \mathcal{C} = sgn(\mathcal{C}_{nb}) = \begin{cases}
 -1\ \ \text{if $\mathcal{C}_{nb}$ < 0}\\
+1\ \ \text{otherwise.} 
\end{cases}
\end{equation}

For each sample $x$, the true label ${y}$ is \textit{one-hot} encoded at the output layer. During the training, the \textit{softmax} function is applied to the output, and the \textit{cross entropy} is used as the training loss function. Thus, the loss function of the output $\textbf{o} = En(\mathbf{X})\times \mathcal{C}$ can be denoted as 
\begin{equation}
    \text{Loss} = \text{CrossEntropy}(\text{softmax}(\textbf{o}), {y})
\end{equation}

Besides, \textit{weight decay} is also an important step in BNN training. Weight decay usually behaves as a $L2$-norm penalty to prevent the weights from evolving too large, which is an effective strategy to mitigate over-fitting during the training. Combined with small gradients accumulated on non-binary class hypervectors, weight decay makes  $\mathcal{C}_{nb}$ more sensitive to the input patterns and less dependent on the weight initialization. Hence, the final empirical loss is given as 
\begin{equation}
    \mathcal{L} = \sum_{i}\text{CrossEntropy}(\text{softmax}(En({x}_i)^T \mathcal{C}), {y}_i) + \frac{\lambda}{2} \left \| \mathcal{C}_{nb} \right \| ^2
\end{equation}
where $({x}_i, y_i)$ is the training sample $i$ and $\lambda$ is a regularization weight. As the training configuration, $Adam$ is selected as the optimizer. Regarding the evaluations in \cite{liu2021adam}, $Adam$ can outperform other $SGD$-based algorithms on the BNN optimization. 

Moreover,  the \textit{dropout} strategy also plays an indispensable role in the equivalent single-layer BNN training. Since updating all weights is likely to  introduces over-fitting, dropout is proposed to greatly prevent the over-fitting  \cite{srivastava2013improving}. Here, despite that \lehdc only has one layer without a complex architecture, its width is large and all the $D$ values corresponding to each class hypervector are straightforwardly updated, based on the gradient of loss. This may force the class hypervectors adapt to the training samples, leading to over-fitting. Hence, dropout is necessary to obtain the better performance on the HDC classification.

With the equivalent BNN model, we propose \lehdc for binary HDC classification. Our training method solves the mentioned limitations in current HDC training strategies and mitigates the over-fitting issue in a principled manner, providing better generalization ability. 
The cross entropy function along with the weight decay and dropout strategies are only used for training the equivalent BNN. After training, the weight matrix $\mathcal{C} = sgn(\mathcal{C}_{nb})$ can be directly used as the class hypervectors. The 
HDC inference process remains the unchanged,  without requiring extra resources. Hence, \lehdc\ induces zero resource and time overhead during inference. Further, our method is inspired by modern BNN training techniques, which have theoretical support to approach the optimum of HDC training, rather than using heuristic training strategies in the existing HDC models.

\begin{table*}[h]

 \captionsetup{width=\linewidth}
    \caption{Inference accuracy (\%) comparison between LeHDC and other strategies. Data are shown with format $mean^{\pm std}$.}
    
    \centering
    \begin{tabular}{lccccccc}
    \toprule
     & MNIST & Fashion-MNIST & CIFAR-10 & UCIHAR & ISOLET & PAMAP & \textbf{Avg Increment} \\
     \midrule
     Baseline Binary HDC &  $80.36^{\pm 0.11}$ & $68.04^{\pm 0.17}$ & $29.55^{\pm 0.35}$ & $82.46^{\pm 0.11}$ & $87.42^{\pm 0.15}$ & $77.66^{\pm 0.01}$ & $-$ \\

    Multi-Model \cite{imani2019searchd} & $84.43^{\pm 0.5}$ & $74.05^{\pm 0.5}$ & $22.66^{\pm 0.59}$ & $82.31^{\pm 0.89}$ & $83.47^{\pm 0.43}$ & $91.87^{\pm 0.85}$ & $+2.22$  \\
    Retraining \cite{imani2019quanthd} & $89.28^{\pm 0.07}$ & $80.26^{\pm 0.27}$ & $28.42^{\pm 1.46}$ & $91.25^{\pm 0.21}$ & $92.70^{\pm 0.12}$ & $95.64^{\pm 0.03}$ & $+8.67$ \\
    \midrule
     \textbf{LeHDC} & $\mathbf{ 94.74^{\pm 0.18}}$ & $\mathbf{87.11^{\pm 0.08}}$ & $\mathbf{46.10^{\pm 0.20}}$ & $\mathbf{95.23^{\pm 0.16}}$ & $\mathbf{94.89^{\pm 0.17}}$ & $\mathbf{99.55^{\pm 0.05}}$ & \textbf{+15.32}\\
    \bottomrule
    \end{tabular}
    
    \label{tab:framecomparison}
\end{table*}

\section{Experiments}
\label{sec:validation}
We evaluate the proposed \lehdc on several selected benchmarks: CV classification tasks (MNIST \cite{726791}, Fashion-MNIST \cite{xiao2017/online}, and CIFAR-10 \cite{cifar10}) and datasets used in the original retraining work \cite{imani2019quanthd} (UCIHAR \cite{ucihar}, ISOLET \cite{isolet}, and PAMAP \cite{pamap}). Our goal is to highlight the advantages of \ouralg over the existing HDC training processes, and hence we mainly compare \ouralg against the existing HDC models. Note that the pros and cons between a general HDC and conventional machine learning models have been extensively studied in the literature, which is thus not the focus in this work \cite{imani2019framework}.

Unless otherwise stated, we adopt the follow configurations in our evaluation.\footnote{Since the existing HDC models in \cite{imani2019quanthd,imani2019searchd} are not open-sourced, we build the retraining and multi-model HDC frameworks by ourselves, and the actual numerical values might differ.} For the retraining strategy, the learning rate is $\alpha = 0.05$, and $\alpha = 1.5$ in the first iteration. We run 150 iterations to ensure the retraining has converged. For the multi-model strategy, we follow the approach in \cite{imani2019searchd} and choose 64 hypervectors per class. For our proposed BNN-training strategy, the hyper-parameters are shown in Table \ref{tab:parameter}. 
As a baseline reference, we test the benchmarks on binary HDC without any retraining.
All the experiments are evaluated with Python on an 3.60GHz Intel i7-9700K CPU with 16GB memory and Tesla P100 GPU with 16GB memory.

\subsection{Model Evaluation}
\begin{figure}[htbp]
\centering
	\subfigure[Training trajectory]{
	    \centering
		\includegraphics[width=0.48\linewidth]{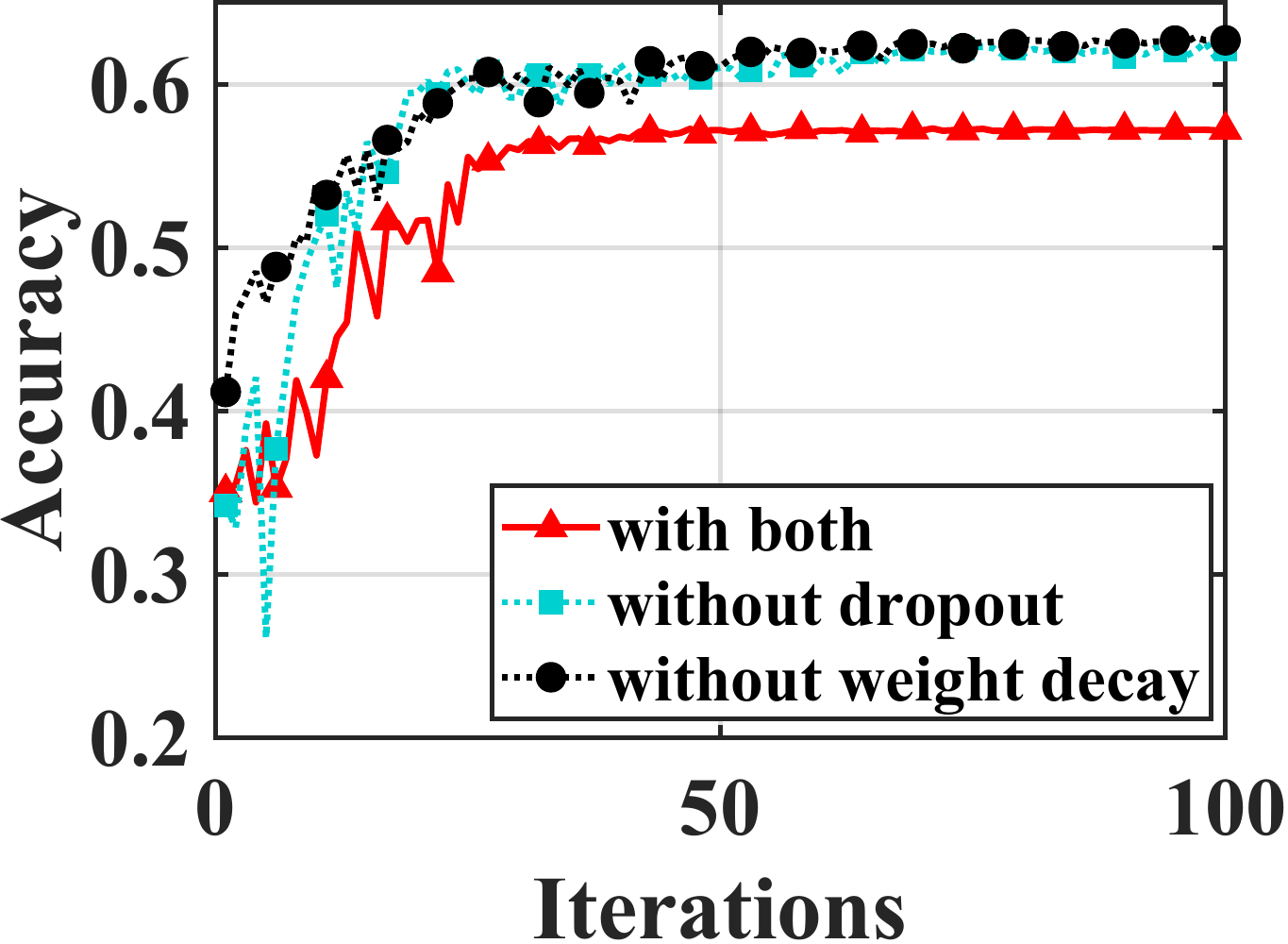}}
	\subfigure[Testing trajectory]{
	    \centering
		\includegraphics[width=0.48\linewidth]{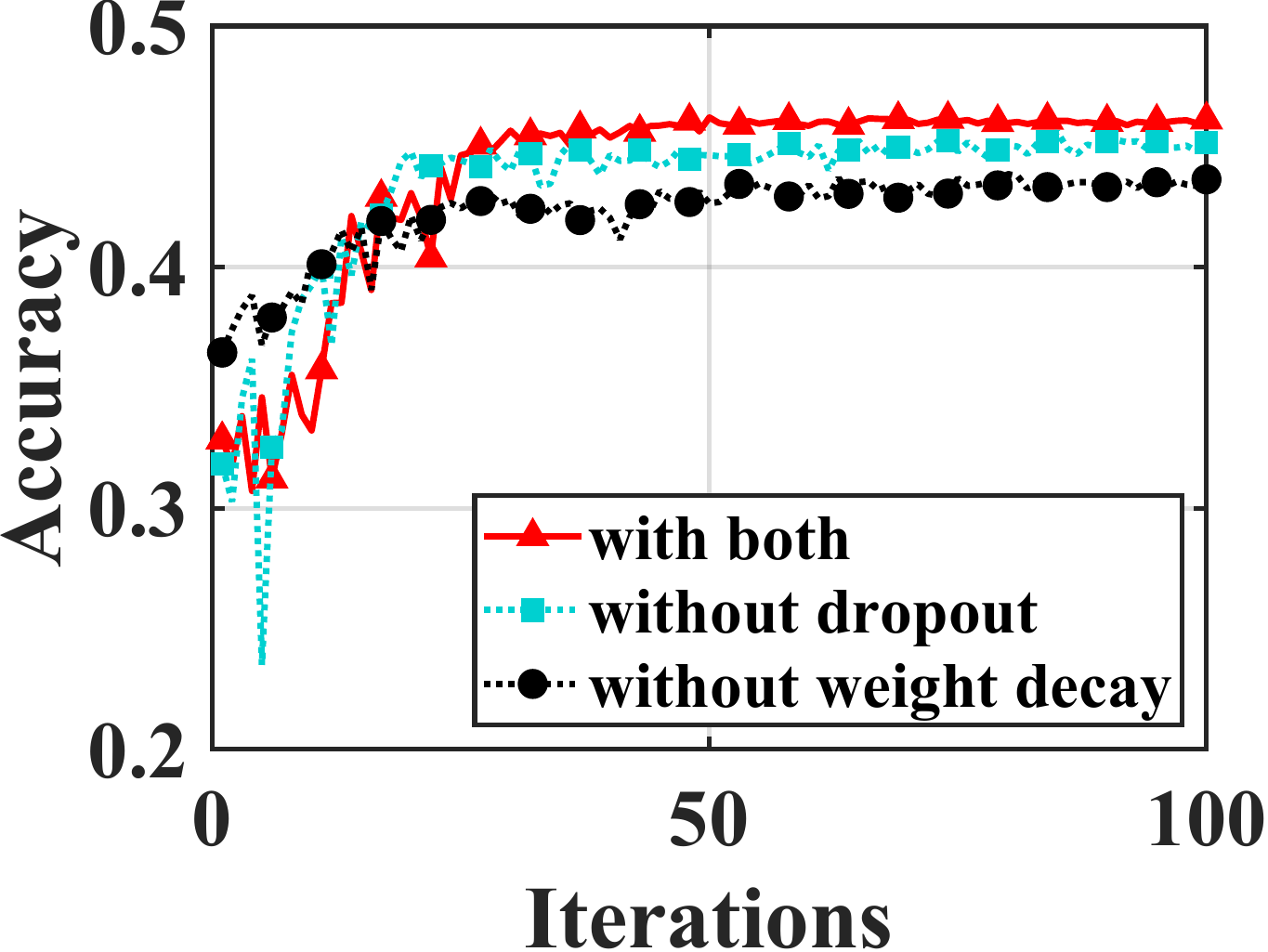}}
\caption{The training and testing accuracy of CIFAR-10 along the iterations. We consider the cases that have weight decay, dropout, and both.}
\label{fig:casestudy}
\end{figure}

First, we validate the significance of weight decay and dropout in \lehdc. In Fig. \ref{fig:casestudy}, we show the training and testing trajectories along the iterations on the CIFAR-10 dataset. By considering the weight decay and dropout during training, the testing accuracy can be increased. An interesting observation is that, if considering both the weight decay and dropout, the training accuracy will decrease. However, the testing accuracy in this case is the highest one. This is due to over-fitting that occurs when either weight decay or dropout is not included. Hence, the trained class hypervectors have better generality. 

\begin{figure}[htbp]
\subfigure[Fashion-MNIST]{
	    \centering
		\includegraphics[width=0.48\linewidth]{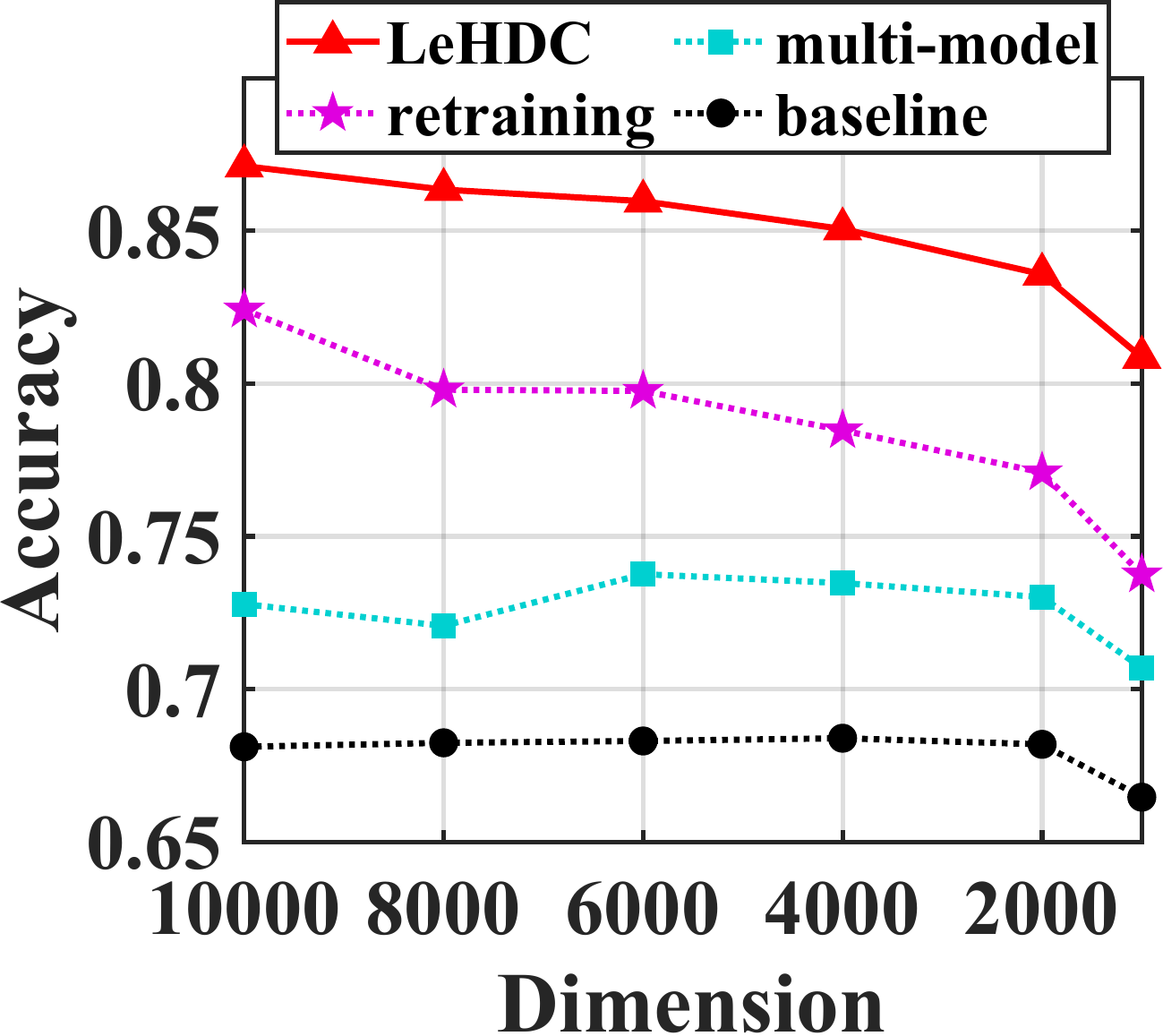}}
		~~
	\subfigure[ISOLET]{
	    \centering
		\includegraphics[width=0.48\linewidth]{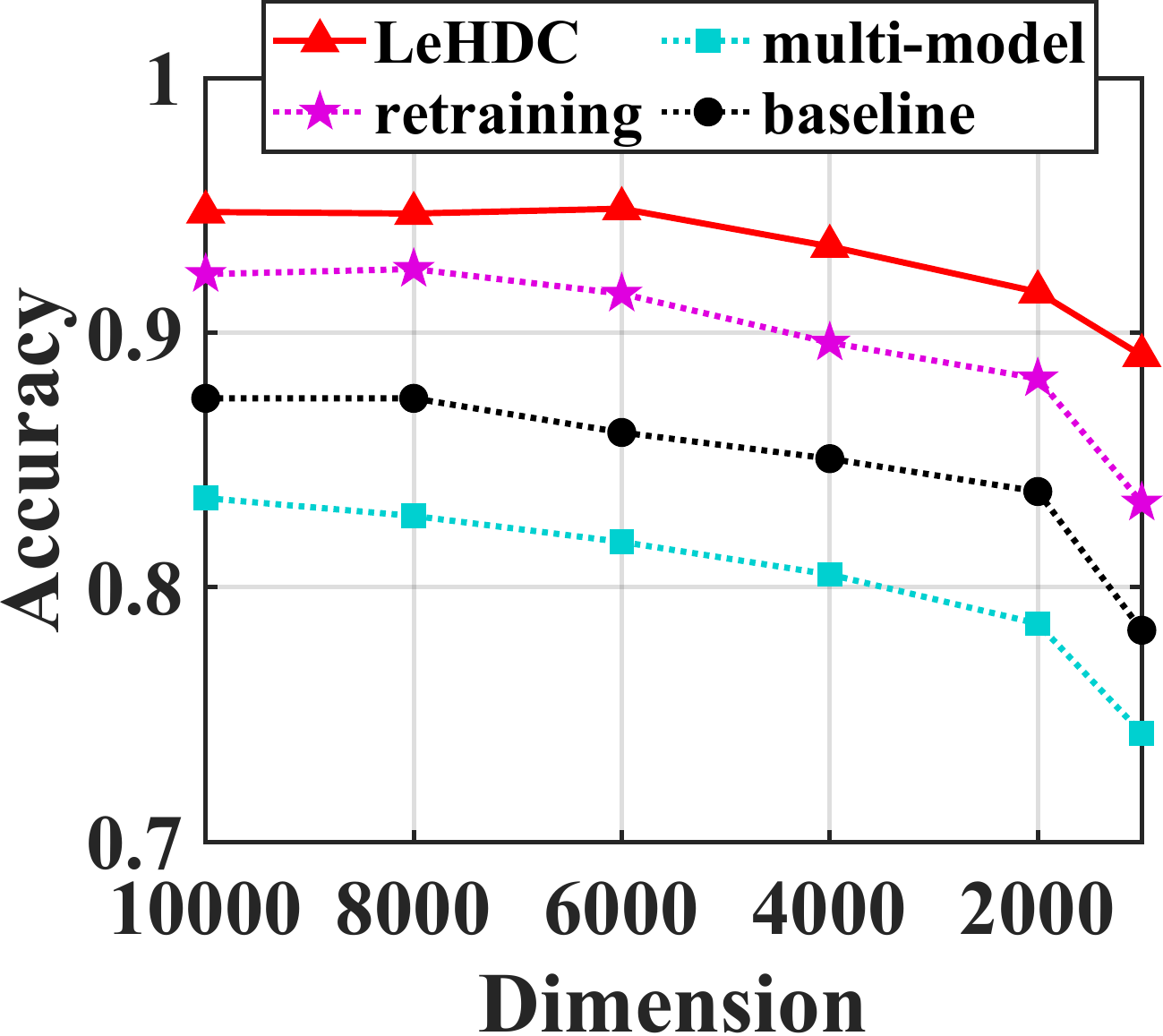}}
\caption{The change of inference accuracy along with the dimension reduction on Fashion-MNIST and ISOLET datasets.}
\label{fig:scalability}
\end{figure}

Further, we evaluate the scalability of \lehdc. We show the accuracy degradation along the dimension reduction across different training strategies in Fig. \ref{fig:scalability}. We can see that \lehdc\ always outperforms other training strategies. Additionally, it achieves the same accuracy as $D=2,000$ as the retraining strategy with a much higher dimension $D=10,000$. 
Another observation is that the multi-model strategy sometimes may even perform worse than the baseline binary HDC, such as on the ISOLET dataset.

Moreover, we discuss the computational resource required for different binary HDC frameworks. Since \lehdc\ only optimizes the training procedure, without inducing extra computation during inference, it has the same time consumption and resource occupation as the baseline and retraining binary HDC. However, multi-model strategy costs more storage due to the multiple class hypervectors. 
Also, hardware acceleration on FPGA and in-memory computing is explored to support the inference in microseconds \cite{imani2019quanthd, imani2019searchd}. Thus,
 \ouralg improves the accuracy performance, with the same energy, latency, and size during inference.

\subsection{Accuracy Improvement}
We evaluate the accuracy performance of \lehdc\ with other HDC training strategies. We fine-tune the training configuration for each dataset, as shown in Table \ref{tab:parameter}. Note that we still use $D=10,000$ for the evaluation, in order to make the comparison fair with other strategies. The learning rate will decay during the training, if the training loss increasing is detected.

\begin{table}[h]
\caption{Hyper-parameters used in \lehdc\ configurations.}
\resizebox{\linewidth}{!}{%
\begin{tabular}{lccccc}
\toprule
\multirow{2}{*}{Dataset} & \multicolumn{5}{c}{Parameters} \\\cline{2-6}
& \textit{WD}$^{\mathrm{1}}$ & \textit{LR}$^{\mathrm{2}}$ & \textit{B}$^{\mathrm{3}}$ & \textit{DR}$^{\mathrm{4}}$ & \textit{Epochs}\\
\hline
MNIST &  $0.05$ & $0.01$ & $64$  &  $0.5$ &  $100$\\
Fashion-MNIST & $0.03$ & $0.1$ & $256$ & $0.3$ & $200$\\
CIAFR-10 & $0.03$ & $0.001$ & $512$ & $0.3$ & $200$ \\
 UCIHAR, ISOLET, PAMAP & $0.05$ & $0.01$ & $64$  &  $0.5$ &  $100$\\
\bottomrule
\multicolumn{6}{l}{$^{\mathrm{1}}$\textit{WD} = Weight Decay $^{\mathrm{2}}$\textit{LR} = Learning Rate $^{\mathrm{3}}$\textit{B} = Batch Size $^{\mathrm{4}}$\textit{DR} = Dropout Rate}
\end{tabular}%
}
\label{tab:parameter}
\end{table}

The inference accuracy comparison is shown in Table \ref{tab:framecomparison}. As shown in the results, baseline HDC performs the worst in most benchmarks. However, the multi-model strategy sometimes even performs worse than the baseline HDC, such as on the CIFAR-10 and ISOLET datasets. By observing the characteristic of these benchmarks, we find that these two datasets have a large number of features or classes, but relatively fewer training samples; thus, the multi-model cannot deal with a complicated HDC classification task without sufficient training samples. Meanwhile, the retraining strategy enhances the training procedure, and has good accuracy improvement against the baseline HDC. On the other hand, our proposed \lehdc\ further improves the inference accuracy against the retraining. Hence, \lehdc\ can make the HDC classification closer to an optimum with zero resource and time overhead during inference.

\section{Conclusion and Discussion}
\label{sec:conclusion}
In this work, we investigate the existing limitations in current HDC training strategies and construct an equivalent BNN to the binary HDC model. Accordingly, we propose \lehdc\ to train the class hypervectors on the BNN structure in a principled manner. The evaluation shows that the learning-based BNN strategy can outperform other HDC training strategies, achieving the best close-to-optimal accuracy performance out, while introducing zero resource and time overhead during the HDC inference. 

Despite that we only fine-tune the explicit hyper-parameters, the \lehdc\ strategy outperforms other training strategies on the selected benchmarks. However, we note there are other implicit ones, such as the ratio of validation set and the learning rate decay along the training.
Moreover, since HDC model can be equivalently represented as neural network models, along with the advances in training BNNs, we expect that the HDC model performance can be further improved by training an equivalent BNN.

Although significantly improving the inference accuracy with the same energy consumption and latency, we admit that the HDC-based inference is still not as powerful as modern DNN framework. For example, a Convolutional Neural Network (CNN) can easily achieve over 90\% accuracy on CIFAR-10. This is mainly due to the fundamental limitations of the existing HDC framework, which is essentially a simple single-layer BNN. 

\input{ref.bbl}

\end{document}